\documentclass[conference]{IEEEtran}
\IEEEoverridecommandlockouts
\usepackage{amsmath,amssymb,amsfonts}
\usepackage{algorithmic}
\usepackage{graphicx}
\usepackage{textcomp}
\usepackage{xcolor}
\usepackage{subfig}  
\usepackage[T1]{fontenc}
\usepackage{caption}
\usepackage{url}
\usepackage[backend=biber,style=ieee]{biblatex}
\def\BibTeX{{\rm B\kern-.05em{\sc i\kern-.025em b}\kern-.08em
    T\kern-.1667em\lower.7ex\hbox{E}\kern-.125emX}}

\newcommand{\linebreakand}{%
  \end{@IEEEauthorhalign}
  \hfill\mbox{}\par
  \mbox{}\hfill\begin{@IEEEauthorhalign}
}
\makeatother

\begin{document}

\title{A desktop learning factory for smart and resilient manufacturing based on digital twin and AI}

\author{
\IEEEauthorblockN{Wenxuan Hu}
\IEEEauthorblockA{\textit{Sorbonne-Université, France.} \\
E-mail: wenxuan.hu@etu.sorbonne-universite.fr}

\and

\IEEEauthorblockN{Zhuoxuan Cao}
\IEEEauthorblockA{\textit{Sorbonne-Université, France.} \\
E-mail: zhuoxuan.cao@etu.sorbonne-universite.fr}
\and
\IEEEauthorblockN{Achraf El Messaoudi}
\IEEEauthorblockA{\textit{Sorbonne-Université, France.} \\
E-mail: achraf.el\_messaoudi@etu.sorbonne-universite.fr}
\and
\IEEEauthorblockN{Peilin Li}
\IEEEauthorblockA{\textit{Sorbonne-Université, France.} \\
E-mail: peilin.li@etu.sorbonne-universite.fr}
\and
\IEEEauthorblockN{Zhiguo Zeng}
\IEEEauthorblockA{\textit{Chair of Risk and Resilience of Complex Systems,} \\
\textit{Laboratoire Génie Industriel, CentraleSupélec, Université Paris-Saclay, France.} \\
E-mail: zhiguo.zeng@centralesupelec.fr}
}

\maketitle
\begin{abstract}
Learning factories have been increasingly applied in engineering research and education. In this paper, we develop a small-scale, low-cost, fully open-sourced learning factory that leverages AI and digital twins to demonstrate smart and resilient manufacturing. The learning factory consists of an AGV, three robots, two conveyors and their digital twins. Three typical use cases are designed to demonstrate the capabilities of the learning factory, i.e., material sorting through computer vision-based robot control, coordinating AGV, robots and conveyors for assembly, and digital twin-driven predictive maintenance. The results show that the developed learning factory can adequately support the teaching and research activities on digital twins and AI on manufacturing. Its fully open-source architecture and relatively low deployment cost make it an ideal solution for universities and research institutions.
\end{abstract}

\begin{IEEEkeywords}
Learning factory, Digital twin, Resilience, Deep learning, Open source
\end{IEEEkeywords}

\section{Introduction}

Learning Factory is an educational platform that closely integrates teaching with the simulation of real production environments \cite{Matt2014}. For example, Free University of Bolzano developed a mini-factory to demonstrate modular reconfigurability through integrated robotics solutions \cite{Matt2014} . EAFIT University's learning factory \cite{Baena2017} adopts comprehensive ICT(Information and Communication Technology) infrastructure to achieve Industry 4.0 compliance, enabling real-time data acquisition and analysis throughout production elements. 

Existing learning factories, as reviewed before, share a few common challenges. First, most existing learning factories use industrial robots and controllers. These industrial equipment is usually expensive, which limits the applicability of the learning factory. Industrial equipment also takes up considerable space, making reconfiguration time-consuming and labor-intensive. Another significant drawback of the learning factories designed with industrial equipment is that, the industrial equipment is usually closed and it is difficult to design and implement new functions freely. Second, most of the existing learning factories still focus on traditional production processes. Emerging technologies such as Artificial Intelligence (AI) and digital twin are not well-integrated in the existing learning factories. Also, predictive maintenance and how to maintain a resilient production system, although important in practice, do not receive sufficient attention in the design of learning factories.

To address these issues, we propose a low-cost, fully open, "desktop-level" learning factory to demonstrate resilient production line operation based on AI and digital twins. The learning factory consists of an AGV, three robots, and two conveyors. The AGV and robotic arms are fully open and built on Robot Operating System (ROS), supported by a user-friendly graphical programming interface. This significantly lowers the entry barrier for beginners while leveraging open-source resources for rapid iteration. Additionally, this solution incorporates deep learning (e.g., image recognition and imitation learning) and digital twin technology, aligning closely with industry 4.0 trends. 

\section{System Architecture}

\begin{figure*}[htb]
    \centering
    \includegraphics[width=.8\linewidth]{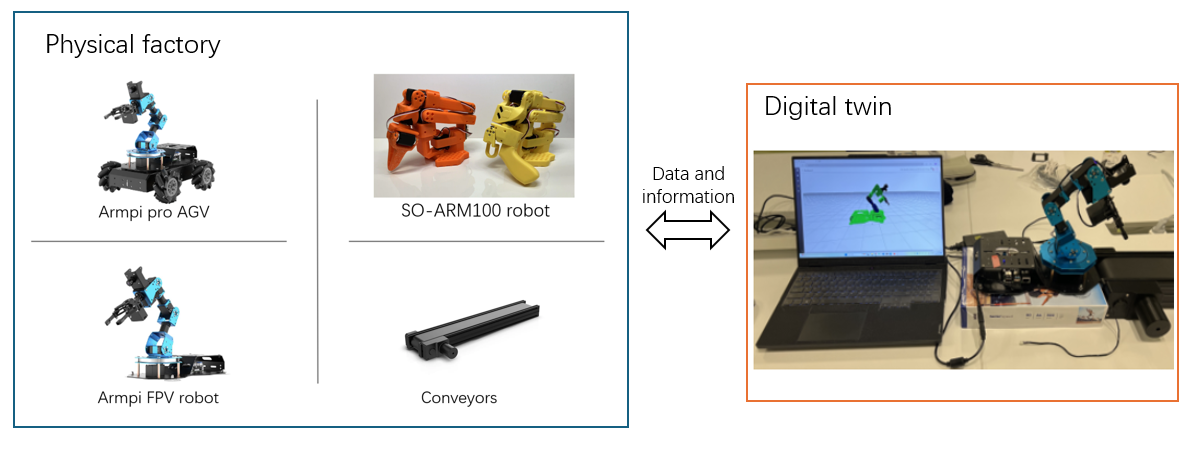}
    \caption{Architecture of the learning factory.}
    \label{fig:sys_architecture}
\end{figure*}

As shown in Figure \ref{fig:sys_architecture}, the developed learning factory mainly comprises four modules: robotic arms for object grasping and assembly (see Sect. \ref{Sect_2.1}) for details), AGV (Automated Guided Vehicle) for flexible material transportation (Sect. \ref{Sect_2.2}), robotic arms based on imitation learning(Sect. \ref{Sect_2.3}) for handling more complex tasks, and digital twin(Sect. \ref{Sect_2.4}) for real-time monitoring and visualization of the production process. The data collected by the digital twin is further analyzed by deep learning models to diagnose faults and optimize scheduling. These modules work closely together to create a flexible, visualized, and continuously optimized production and learning environment.

\subsection{ArmPi FPV Robotic Arm}\label{Sect_2.1}


Robotic arms are used in this learning factory to perform tasks like grasping, sorting, and assembly. In this learning factory, we use two types of robots. In this section, we introduce the first type, the ArmPi FPV from Hiwonder\cite{HiwonderFPV} (see Figure \ref{fig:sys_architecture}). The robotic arm is equipped with an onboard camera that identifies a target object’s relative position in the image frame. The underlying control mechanism relies on inverse kinematics to determine joint angles required for achieving the target posture of the end-effector. Subsequently, precise joint movements are achieved through PID control, ensuring smooth and efficient operations. Robotic arms can transfer materials from conveyor belts to workstations or accurately place and sort finished or semi-finished products. The robot is controlled by a Rapsberry 4B single-board computer that runs ROS 1 Melodic.

\begin{figure*}[htbp]
    \centering
    \includegraphics[width=.75\linewidth]{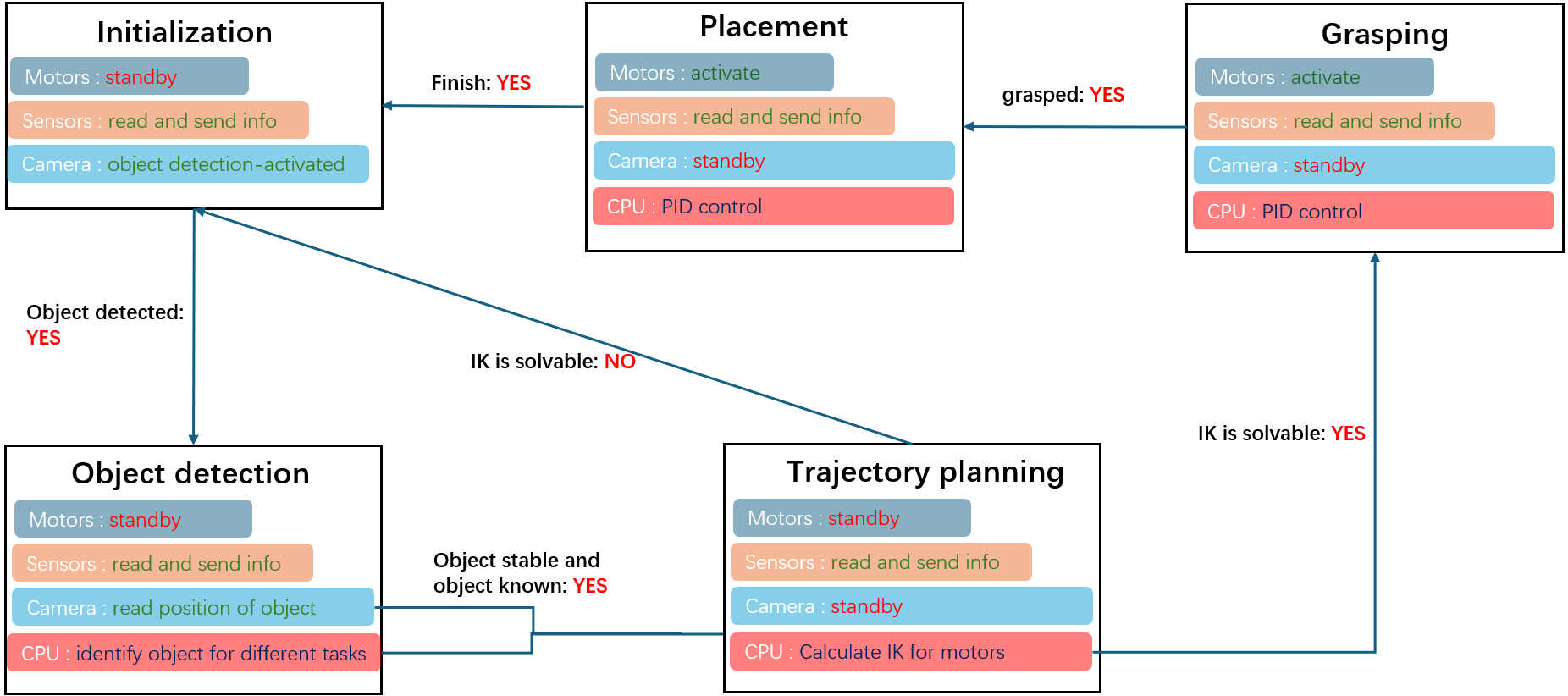}
    \caption{Flowchart of task grasping by Hiwonder ArmPi.}
    \label{fig:Flowchart_task_grasp}
\end{figure*}

A typical task of this robot is to pick up a desired material and move and place it on a desired position. A flowchart of a task implementation process is shown in Figure~\ref{fig:Flowchart_task_grasp}. The task execution process includes initialization, object detection, trajectory planning, grasping, and placement steps.

\subsection{AGV Control and Material Transportation}\label{Sect_2.2}

In this learning factory, we employ the Hiwonder ArmPi Pro AGV \cite{HiwonderPro}, as illustrated in Figure \ref{fig:sys_architecture} to identify and transport required materials to the buffer zone or conveyor belts. By color detection or April-tag, the AGV can recognize target materials and plan the optimal navigation route. The navigation process typically employs linear or nonlinear path detection, combined with closed-loop navigation algorithms, to dynamically adjust speed and direction in real-time, ensuring accurate and stable transportation.

To efficiently control the robot and enable it to complete different tasks in specific zones, we adopted a finite state machine (FSM) as shown in Figure \ref{fig:sys_architecture}. 

\begin{figure}[htbp]
    \centering
    \includegraphics[width=1\linewidth]{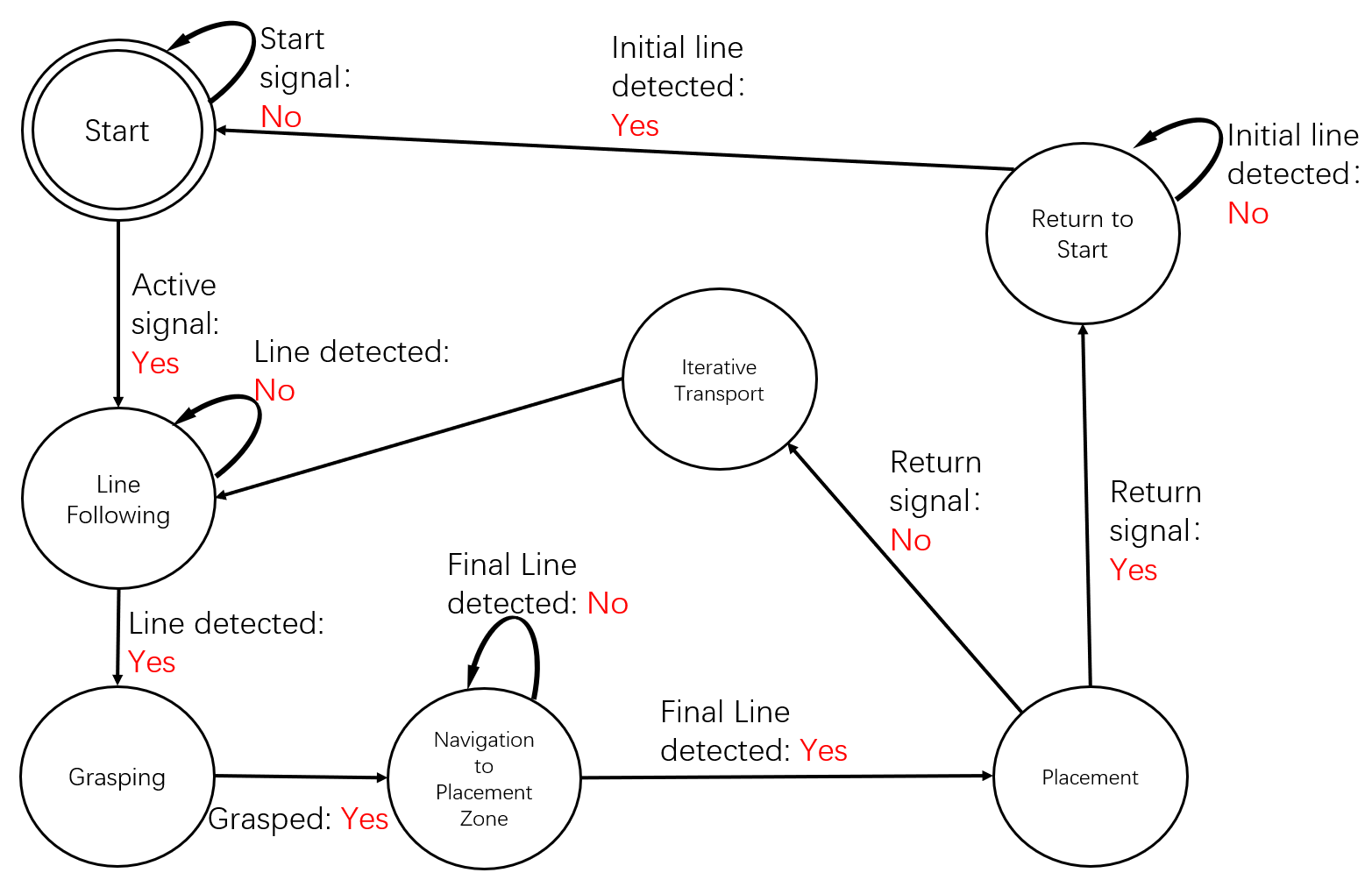} 
    \caption{Finite state machine structure of the AGV.}
    \label{fig:FSM_AGV} 
\end{figure}

The different modes in Figure \ref{fig:sys_architecture} are defined below:
\begin{itemize}

    \item Start Mode:  
    The robotic arm stay at initial position and waiting for start signal.
    
    \item Line Following Mode:
    The robot activates line following mode, using its camera to detect the guiding line on the ground. It adjusts its trajectory in real time to follow this line and move stably to the material zone (collection area).

    \item Grasping Mode: 
    Once in the material zone, the robot switches to grasping mode (by same function shown in previous section \textit{ArmPi FPV Robotic Arm}). It uses its vision system to detect the target object’s position and orientation. The robotic arm then adjusts its posture to execute the grasping.

    \item Navigation to Placement Zone:
    After grasping the object, the robot switches back to line following mode and follows the guiding line to the placement zone.

    \item Placement Mode:
    In the placement zone, the robot switches to placement mode. It adjusts the robotic arm’s position to stably place the object onto the conveyor belt.

    \item Iterative Transport Mode:  
    The robot automatically returns to the material zone and repeats the Grasping $\to$ Navigation $\to$ Placement process until all objects have been transported.

    \item Return to Start Mode: 
    Once all tasks are completed, the robot activates the reverse mode and follows the guiding line back to its starting point, entering standby mode.
\end{itemize}

\begin{figure}[htbp]
    \centering
    \includegraphics[width=0.5\linewidth]{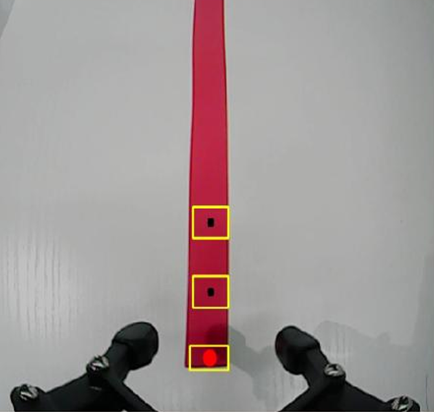} 
    \caption{Line Detection Results}
    \label{fig:AGV_Line_follow} 
\end{figure}

When the AGV operates in the following Mode, we use image processing techniques to identify a guidance line and determine its central coordinates. This guidance line is represented by a red adhesive tape, which serves as a visual reference for the robot’s path planning. This information is then used for trajectory tracking and navigation control. An example of the line detection result is given in Fig. \ref{fig:AGV_Line_follow}. The transitions between modes are based on detecting red lines on the ground that are wider than 10 cm. When a valid marker is detected, the camera analyzes the robot’s current state and determines the next mode to activate. 

\subsection{Robotic arms for imitation learning}\label{Sect_2.3}

For more complex or variable tasks, imitation learning can significantly reduce the programming effort of the robotic arms  \cite{Imitation_learning}. Operators first demonstrate the target operation, such as fine assembly or specific trajectory motion, while the system records and extracts key motion features. Using demonstration data, a model is trained to replicate the actions demonstrated. Once trained, the robotic arm can reproduce the essential aspects of the demonstration and quickly switch or update workflows as needed. In this learning factory, we also include a robot named SO-ARM100 \cite{TheRobotStudio2025} for imitation learning, as shown in Fig. Figure \ref{fig:sys_architecture}.



To support IL on our robot, we implement a Action Chunk Transformer (ACT) \cite{zhao2023learning} model to learn from human demonstrations while adapting them to our 6-joint robotic arm. The \textit{ACT} (as shown in Fig. \ref{fig:detr}) is a Transformer-based architecture designed for continuous action generation from robotic demonstrations. It distinguishes itself by grouping consecutive actions into "chunks" before predicting the next motion sequence. 

To control a robotic arm, the position (or angle) of each joint is specified at regular intervals (e.g., 50 times per second). By recording the complete state of all joints at each instant~$t$, a temporal sequence describing the robot’s movement is obtained. The goal of ACT is to learn to predict future positions over multiple time steps, given an input set (such as the current robot state or a visual representation of the scene).

\subsubsection{Overview of the ACT architecture}

\begin{figure}[htbp]
    \centering
    \includegraphics[width=0.9\linewidth]{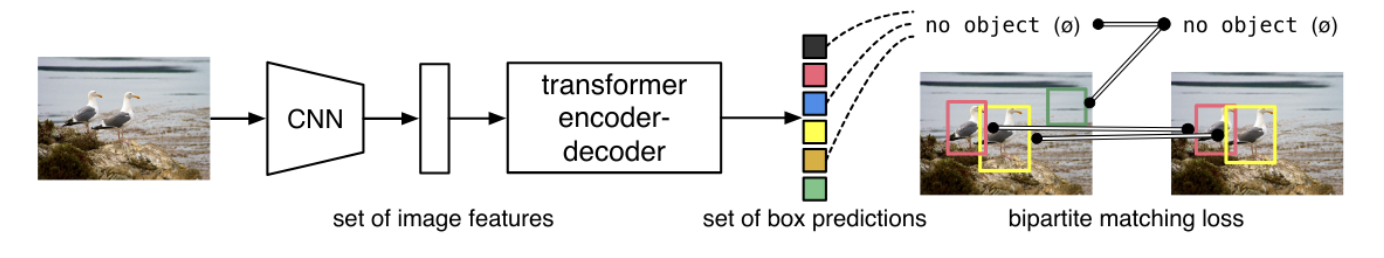}
    \caption{DETR directly predicts (in parallel) the final set of detections by combining a conventional CNN with a Transformer architecture. During training, a bipartite matching uniquely associates predictions with ground truth boxes. A prediction without a match is assigned to the class  no object  ($\emptyset$).}
    \label{fig:detr}
\end{figure}

The ACT adopted in this project follows an Encoder-Decoder approach similar to \textit{DETR} (\textit{DEtection TRansformer}) \cite{DETR}, but used for action generation rather than object detection(as shown in Fig. \ref{fig:detr}):
\begin{itemize}
    \item Encoder: Processes the current state of the robot (and optionally other modalities like images) and produces \textit{embeddings} rich in information about the current configuration.
    \item Decoder: Instead of textual tokens, \textit{queries} are used for each action to be predicted in the chunk. The decoder transforms these queries into successive joint positions. Unlike traditional Transformers, no \textit{causal masking} is applied: future actions within a chunk can influence each other.
\end{itemize}

\subsubsection{Adapting ACT to a 6-DOF Robot}
The original ACT targeted a 14-dimensional dual-arm system (7 DoF per arm). For our project, we adapted it to a single 6-joint arm. Key modifications included:
\begin{itemize}
    \item Reducing input and output dimensions from 14 to 6, corresponding to our 6 control axes.
    \item Adjusting fixed model parameters referring explicitly to 14 dimensions.
    \item Recalibrating the latent variable \( z \): we reduced its dimension to fit the lower complexity of a single arm.
\end{itemize}
These changes were relatively simple thanks to the modular structure of the code and the flexibility of the Transformer architecture.

\subsubsection{Training by Demonstration and Generation}
The training process follows these steps:
\begin{enumerate}
    \item Data collection: The trajectory of the robotic arm (joint angles) is recorded at a fixed frequency while a person \textit{demonstrates} a movement (e.g., 5 seconds at 50\,Hz).
    \item Chunk construction: Each sequence of $n$ consecutive positions serves as the target that the decoder must produce, while the state at instant $t$ serves as input to the encoder.
    \item Loss function: The decoder's predicted (chunk) sequence is compared with the actual sequence using a loss function such as $L_1$ or MSE.
    \item CVAE training: Simultaneously, the latent variable \( z \) learns to encode the \textit{shape} of the trajectory via a variational auto-encoder mechanism, enabling the model to handle diverse movement styles.
\end{enumerate}
During inference, the model generates predictions \textbf{chunk by chunk}, allowing the robotic arm to anticipate multiple time steps in advance.

In this project, we apply imitation learning to the packaging stage of a robotic arm, demonstrating remarkable flexibility and efficiency. Specifically, with only five demonstration episodes and approximately 20 epochs of training, the robotic arm can reliably perform relatively complex packaging tasks within about 1.5 hours in a Google Colab environment equipped with an H100 GPU. Compared to traditional manual programming methods, this imitation learning-based approach not only adapts to various packaging forms but also allows multiple policies to be saved and switched between different tasks, significantly enhancing the scalability and creativity of the system in learning factory scenarios.

\subsection{\textbf{Digital Twin and Fault Diagnosis}}\label{Sect_2.4}

The digital twin module (shown in Fig. \ref{fig:sys_architecture}) mirrors the real-time status of physical devices such as robotic arms and AGVs, dynamically displaying the production process through a visualization interface. Meanwhile, the system conducts multidimensional analysis on accumulated process data and utilizes deep learning models, such as LSTM, for fault diagnosis. After preprocessing the data, these models capture potential anomalies or failure trends in time-series data, providing predictive maintenance insights and production optimization strategies. This approach enhances system resilience and ensures continuous improvement.

The digital twin system consists of four main modules:  
Robot, Web Server, Backend, and Frontend.  
It is based on ROS, WebSocket, and a machine learning model to visualize, collect, and analyze data.

\begin{figure}[htbp]
    \centering
    \includegraphics[width=0.9\linewidth]{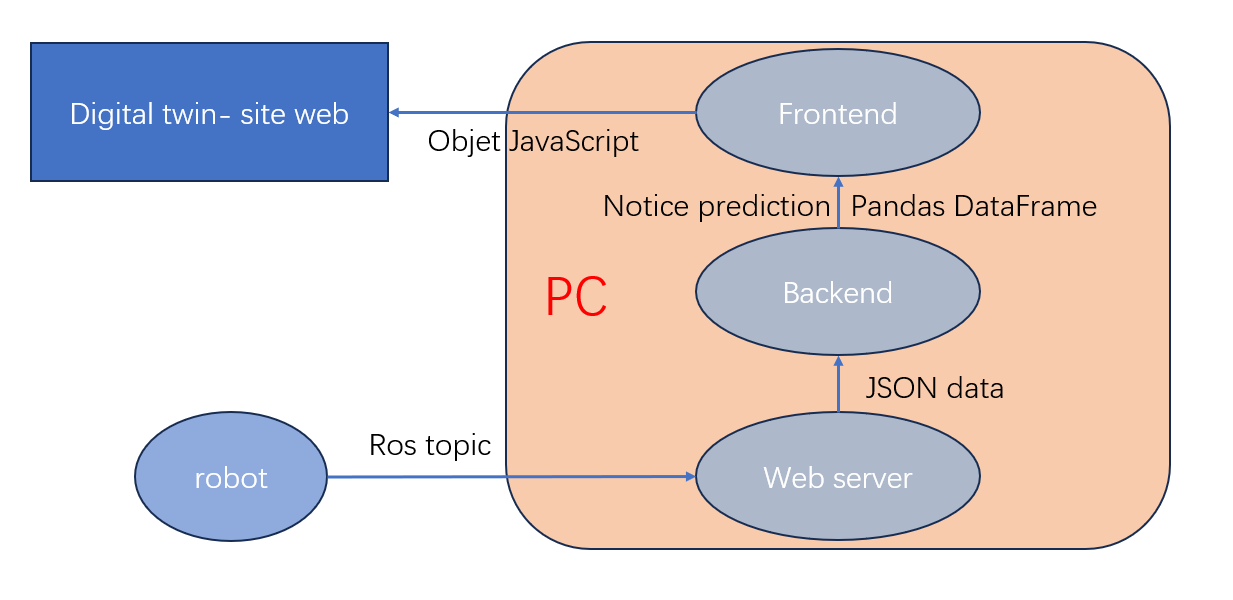}
    \caption{Communication architecture of the digital twin}
    \label{fig:structure_DT}
\end{figure}

As shown in Fig. \ref{fig:structure_DT}, the robot runs a ROS node to monitor motor status and collect position, temperature, and voltage data. This information is published as a \texttt{rostopic} at a frequency of 10Hz and transmitted to the PC. The web server subscribes to this \texttt{rostopic} based on the IP address, following the ROS master-slave structure. It converts the \texttt{rostopic} data into JSON format for transmission via WebSocket and provides two connection modes: in the first mode, the robot acts as a WiFi access point, allowing the PC to connect directly, which is suitable for autonomous systems; in the second mode, both the robot and PC connect to the same WiFi network via a router, enabling more flexible and multi-device communication.

On the backend, the received JSON data is converted into a Pandas DataFrame for storage and processing. A pre-trained prediction model based on motor temperature is employed to diagnose potential faults in the robotic arm, identify the failing motor, and generate corresponding alerts. The diagnostic results and real-time data are then transmitted to the frontend. The frontend, developed using Vue.js, provides an interactive interface to monitor the digital twin’s data in real time. It connects to the backend via WebSocket, continuously displaying motor data while also offering a feature for storing and downloading operational history for future analysis.

By integrating these four core modules, the system can flexibly adapt to various production scenarios while maintaining strong scalability and upgrade potential. Real-time data acquisition and analysis enable proactive intervention before failures occur. Combined with IL, the system can quickly respond to process adjustments and new task deployments, offering an innovative and practical solution for both educational and industrial applications.

\section{Use Cases}

In this project, we designed and implemented three typical use cases based on the learning factory: material sorting, assembly, and real-time monitoring and fault detection using digital twin technology. 

For the sorting task, we use color recognition as an example, where the robotic arm can classify materials based on different colors. In the assembly task, to achieve higher stability and precision, we employ AprilTag for identifying and positioning target objects. Meanwhile, leveraging digital twin technology, the entire operation process is visualized and monitored in real time, facilitating production tracking and providing data support for subsequent fault diagnosis and system optimization. When equipment conditions permit, these two stages can be integrated into a complete production line: first, the AGV and robotic arm perform color-based sorting; then, the sorted materials are transported to the packaging workstation, where AprilTag enables high-reliability assembly and sealing. Finally, real-time monitoring through the digital twin ensures centralized management of the entire production process. This modular and scalable design provides a flexible and efficient operational model for the learning factory, laying a solid foundation for further teaching and research advancements.

\subsection{Sorting Task}

\begin{figure}[htbp]
    \centering
    \includegraphics[width=0.9\linewidth]{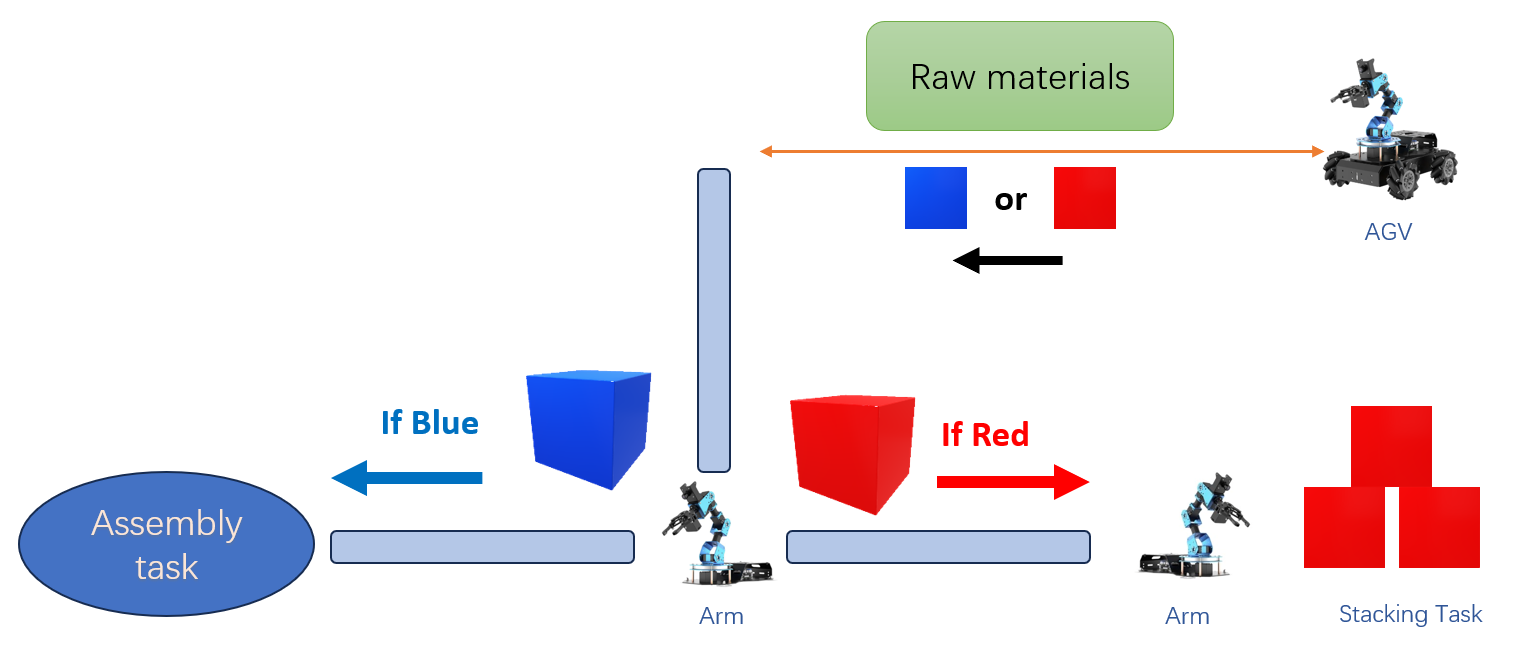}
    \caption{Material sorting and stacking Task}
    \label{fig:sorting}
\end{figure}

In the "Sorting" scenario(Figure~\ref{fig:sorting}), the AGV retrieves raw materials (e.g., blue or red blocks) from the material area based on instructions and places them onto the conveyor belt. The robotic arm then utilizes vision detection or label recognition to classify objects based on characteristics such as color or shape, subsequently sorting them into designated areas or bins. In this example, if the material is a blue block, it is transferred to the next assembly process, whereas red blocks are stacked or piled up. This process not only reduces manual intervention but also enables flexible adaptation to different color or attribute requirements, demonstrating the high configurability of the desktop-level learning factory system.

\subsection{Assembly Task}

\begin{figure}[htbp]
    \centering
    \includegraphics[width=0.9\linewidth]{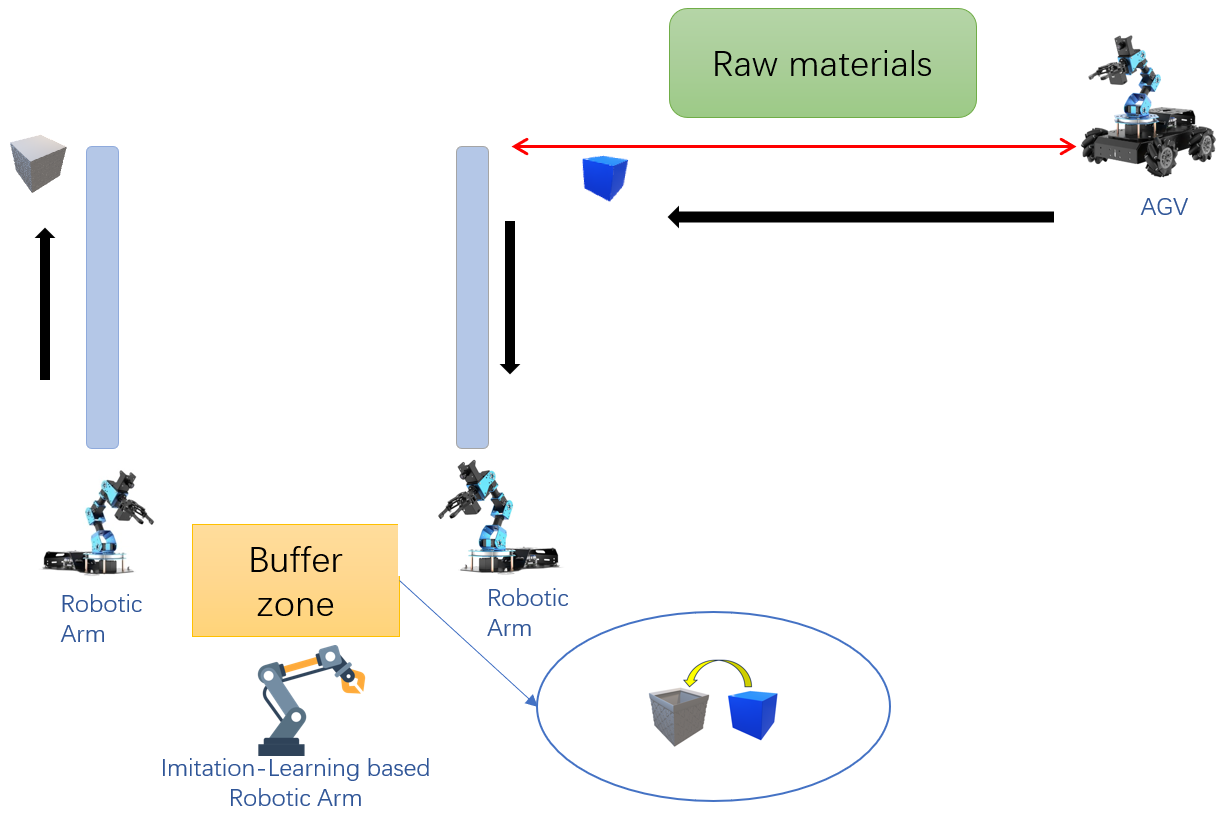}
    \caption{Material assembly Task}
    \label{fig:assembly}
\end{figure}

\begin{figure}[htbp]
    \centering
    \includegraphics[width=0.5\linewidth]{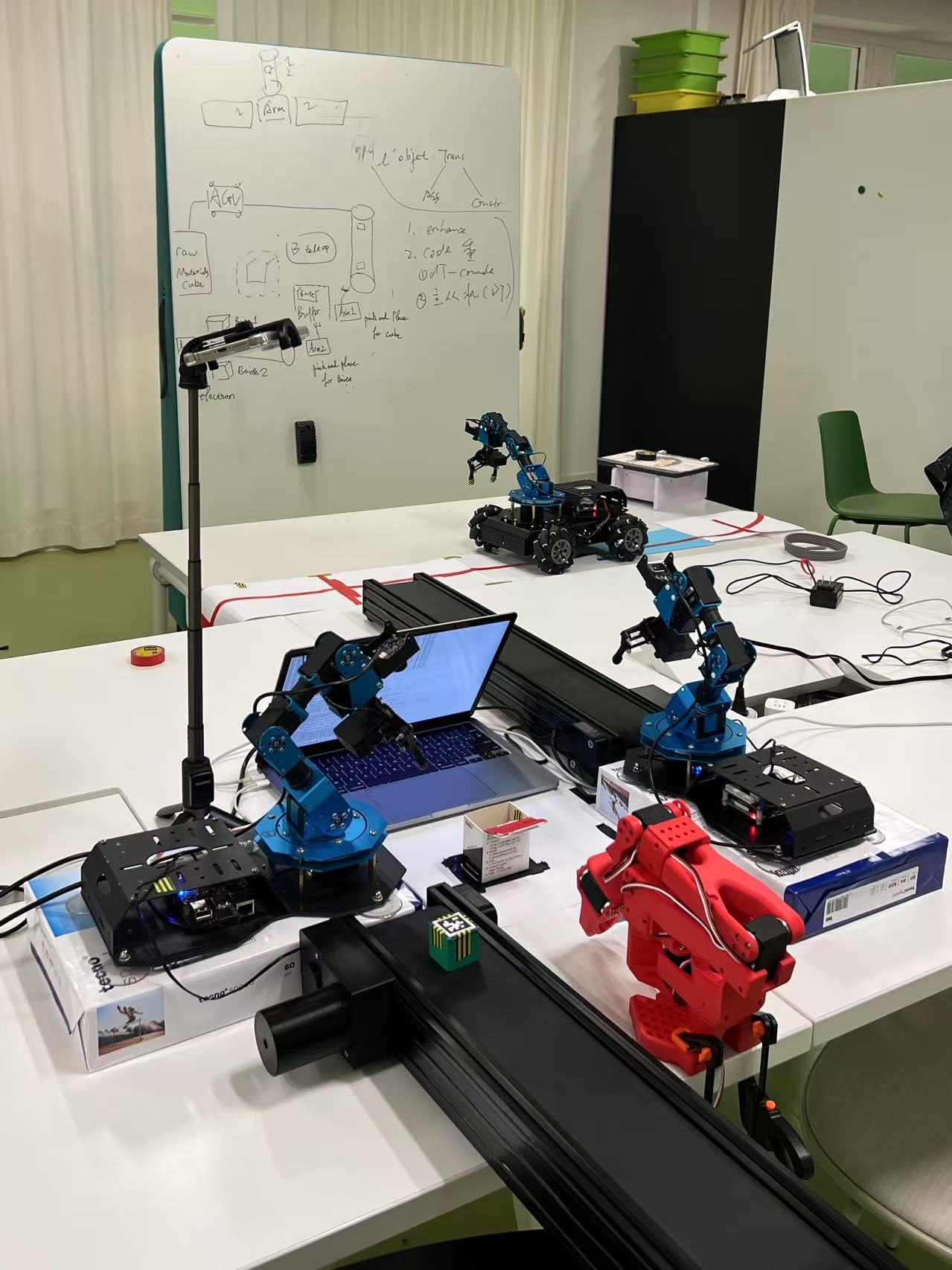}
    \caption{Physical assembly task setup within the desktop learning factory}
    \label{fig:Pic_assembly}
\end{figure}

In this case, the AGV and robotic arms transport the materials to the workstation, where another imitation learning-based robotic arm performs collaborative packaging or component assembly(Figure~\ref{fig:assembly}) . Under helps of IL robotic arm, the sorted materials undergo further processing or assembly in the workstation before being transferred by the robotic arm or AGV to the next production area or storage location. This ensures a flexible and efficient packaging workflow(Figure~\ref{fig:Pic_assembly}).

\subsection{Digital Twin Real-Time Monitoring and Fault Detection}
\begin{figure}[htbp]
\centering
\includegraphics[width=0.9\linewidth]{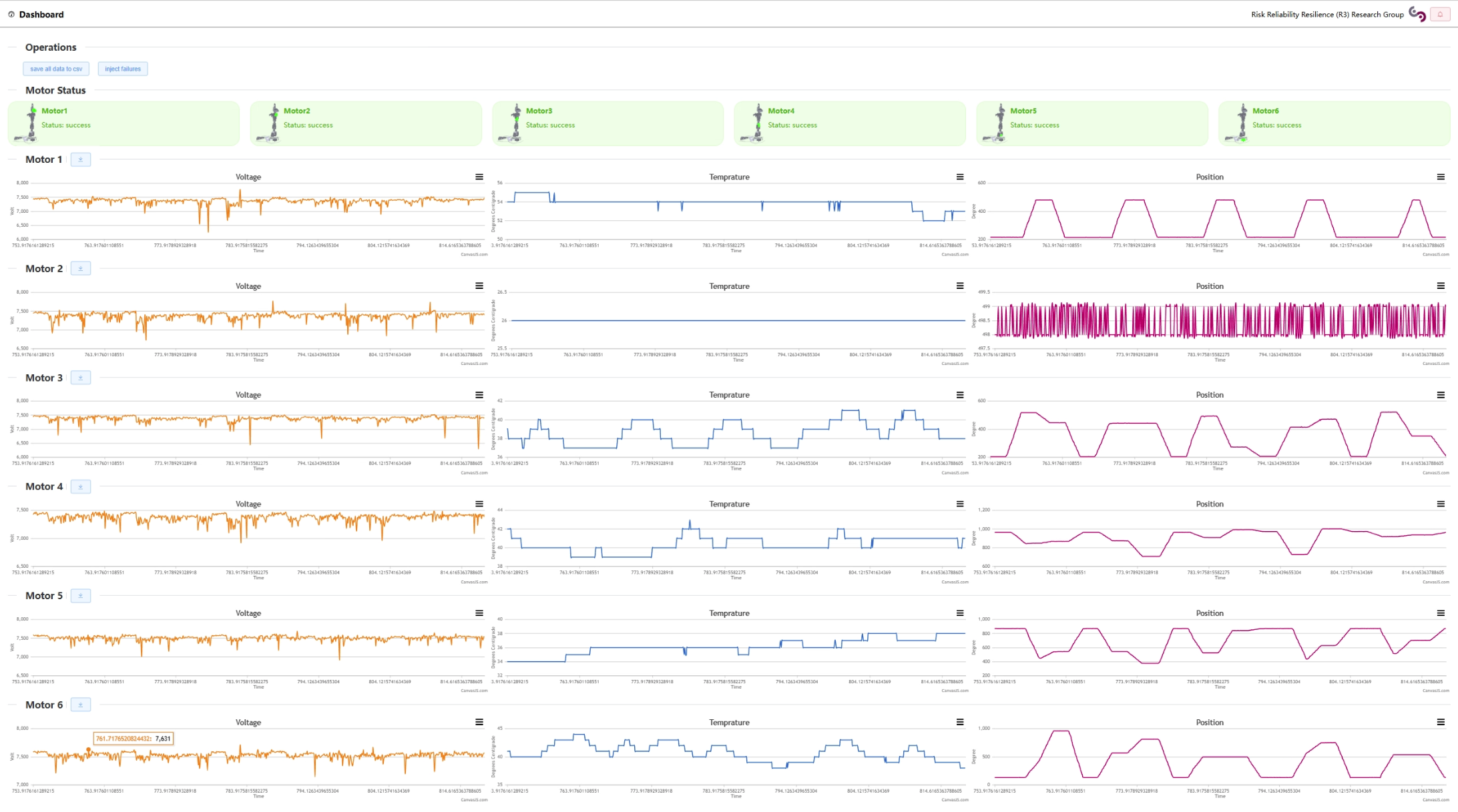}
\caption{Real-time monitoring of six robotic motors in the digital twin interface, showing voltage, temperature, and position traces for each motor}
\label{fig:DT-data}
\end{figure}

To enhance the visualization (Figure~\ref{fig:DT-data}) and traceability of the sorting and packaging processes, the system integrates a digital twin module (previously illustrated in the “Digital Twin” diagram). As the robotic arm and AGV execute their tasks, ROS messages are transmitted in real time to a PC, where the backend processes these data before forwarding them to the web-based frontend for dynamic rendering. The digital twin interface synchronously displays joint movements of the robotic arm, material locations, and fault prediction information. Whenever an anomaly is detected by the AI model within the digital twin, alerts are immediately issued. This mechanism allows operators to oversee and manage the production flow both locally and remotely, significantly bolstering the system’s robustness and maintainability.

In addition, the system continuously collects and analyzes historical error and warning logs to identify common fault patterns and risk factors, thereby refining the robotic arm’s programming logic. For anomalies frequently occurring under specific operating conditions, imitation learning enables the robotic arm to adapt on the fly and autonomously learn to mitigate potential failures. The combination of real-time detection and offline analytics broadens the scope of resilience and reliability improvements for the learning factory. This integrated approach empowers personnel to promptly devise response strategies when encountering production fluctuations or uncertainties, guiding the learning factory toward greater intelligence and adaptability.

\section{Conclusion}
In this paper, we presented a desktop-level learning factory that integrates open source robotics, AI-driven modeling, a digital twin framework, and resilient scheduling strategies to address the challenges of modern manufacturing education. Compared to conventional Learning Factory implementations, our approach emphasizes real-time adaptability by merging data analytics, fault diagnosis, and dynamic production scheduling within a lightweight, reconfigurable setup. By leveraging open-source robotic platforms, deep learning algorithms, and a comprehensive digital twin visualization, students and practitioners can gain hands-on experience in orchestrating resilient operations—ranging from automated sorting and assembly to proactive fault detection and production adjustments. This holistic integration of AI and digital twin technologies not only fosters a deeper understanding of Industry 4.0 concepts, but also equips learners with the skills required to design, monitor, and optimize complex manufacturing systems under rapidly changing conditions. Ultimately, our work offers an accessible yet powerful solution for educational institutions and industries seeking to advance resilient manufacturing practices and cultivate future-ready talent. This learning factory can be used to support research related to digital twins of manufacturing systems, AI in manufacturing and predictive maintenance and operation planning based on digital twins. It can also be used to support teaching on Bachelor's and Master's level in areas like control engineering, AI, manufacturing and operation research.

\section*{Acknowledgment}

This project is supported by the French Research Council under contract number (ANR-22-CE10-0004). The research of Zhiguo Zeng is supported by Chair on Risk and Resilience of Complex Systmes (Chair EDF, Orange and SNCF).






\printbibliography

\end{document}